\documentclass{article}

\PassOptionsToPackage{numbers}{natbib}

\usepackage[preprint]{neurips_2025}

\usepackage[utf8]{inputenc} 
\usepackage[T1]{fontenc}    
\usepackage{hyperref}       
\usepackage{url}            
\usepackage{booktabs}       
\usepackage{amsfonts}       
\usepackage{nicefrac}       
\usepackage{microtype}      
\usepackage{wrapfig}
\usepackage{graphicx}

\usepackage{amsmath}
\usepackage[capitalise]{cleveref}
\usepackage{bm}
\usepackage{algorithm}
\usepackage{algorithmic}
\usepackage{amsthm}
\usepackage{subcaption}
\usepackage{threeparttable, float}
\usepackage{array}
\usepackage{multirow}
\usepackage{nicematrix}
\usepackage{rotating}
\newtheorem{theorem}{\bf Theorem}

\newtheorem{assumption}{\bf Assumption}

\usepackage{nicematrix} 
\usepackage{multirow}   

\definecolor{Gray}{gray}{0.85}

\newcommand{\compilehidecomments}{false}
\ifthenelse{ \equal{\compilehidecomments}{true}}{
    \newcommand{\jinhang}[1]{}
    \newcommand{\rev}[1]{}
}{
    
    \newcommand{\rev}[1]{{\color{red}#1}}

}

\usepackage{lipsum} 

\usepackage[most]{tcolorbox}
\usepackage{lipsum} 
\usepackage{listings}
\tcbuselibrary{breakable, skins, listings, theorems}

\title{Online Multi-LLM Selection via Contextual Bandits under Unstructured Context Evolution}

\author{Manhin Poon$^{1}$ \quad Xiangxiang Dai$^{2}$ \quad Xutong Liu$^{3}$ \quad 
Fang Kong$^{4}$ \\ \textbf{John C.S. Lui}$^{2}$ \quad \textbf{Jinhang Zuo}$^{1}$ \\
$^{1}$City University of Hong Kong \quad 
$^{2}$The Chinese University of Hong Kong\\
$^{3}$Carnegie Mellon University \quad 
$^{4}$Southern University of Science and Technology}

\definecolor{bluex}{rgb}{0.27, 0.42, 0.81}
\definecolor{purplex}{HTML}{9564bf}
\definecolor{red3}{HTML}{C52A20}
\definecolor{red2}{HTML}{B36A6F}
\definecolor{red1}{HTML}{FFb5b5}
\definecolor{purple}{HTML}{B36A6F}
\definecolor{darkyellow}{HTML}{D5BA82}
\definecolor{blue1}{HTML}{508AB2}
\definecolor{blue2}{HTML}{C4E4E3}
\definecolor{green1}{HTML}{A1D0C7}
\definecolor{green2}{HTML}{BFF6BA}
\definecolor{green3}{HTML}{028100}
\definecolor{teal}{HTML}{508AB2}
\definecolor{orange3}{HTML}{c28c69}
\definecolor{blue3}{HTML}{3b75af}
\definecolor{White}{rgb}{1.0, 1.0, 1.0}
\definecolor{DarkSlateGray}{rgb}{0.18431, 0.30980, 0.30980}

\newtcolorbox{mybox}{colback=white!5!white,colframe=black!75!black, left=.05in, right=.05in}

\newtcbtheorem{emptybox}{}%
{colback=green2!5,colframe=blue1,fonttitle=\bfseries,theorem style=plain, left=.0in, right=.0in,bottom=.02in, top=.02in,terminator sign none}{emptybox}

\newtcbtheorem[]{emptyboxx}{A Simple Template For Creating Backward Question}%
{colback=green2!5,colframe=blue1,fonttitle=\bfseries,theorem style=plain, left=.0in, right=.0in,bottom=.02in, top=.02in,terminator sign none}{emptyboxx}

\newtcbtheorem{bth}{Theoreme}
{colback=red!20,colframe=red,theorem style=plain,fonttitle=\bfseries,coltitle=black,terminator sign none}{th}

\newtcbtheorem[]{exmp}{Example}%
{colback=green2!5,colframe=blue1,fonttitle=\bfseries, left=.02in, right=.02in,bottom=.02in, top=.02in}{exmp}

\newtcbtheorem[]{exmp1}{Example}%
{colback=green2!5,colframe=blue1,fonttitle=\bfseries, left=.02in, right=.02in,bottom=.02in, top=.02in}{exmp1}

\newtcbtheorem[]{exmp2}{Example}%
{colback=green2!5,colframe=red3,fonttitle=\bfseries, left=.02in, right=.02in,bottom=.02in, top=.02in}{exmp2}

\newtcbtheorem[]{prompt}{Template For Creating Backward Question}%
{colback=green!5,colframe=green!35!black,left=.0in, right=.0in,bottom=.02in, top=.02in,fonttitle=\bfseries}{th}

\newtcbtheorem[number within=section]{thm}{Theorem}%
{colback=green!5,colframe=green!35!black,fonttitle=\bfseries}{th}

\newtcbtheorem[number within=section]{corr}{Corollary}%
{colback=green!5,colframe=green!35!black,fonttitle=\bfseries}{th}

\newtcbtheorem{method}{Method}%
{colback=green!5,colframe=green!35!black,fonttitle=\bfseries}{th}

\newtcbtheorem{ebox}{}
{colback=green2!5,colframe=blue1, left=.02in, right=.02in,bottom=.02in,title empty, top=.02in, theorem style=plain apart,frame empty}{ebox}
\PassOptionsToPackage{svgnames,dvipsnames,table}{xcolor}
\begin{document}
\maketitle

\begin{abstract}
Large language models (LLMs) exhibit diverse response behaviors, costs, and strengths, making it challenging to select the most suitable LLM for a given user query. We study the problem of adaptive multi-LLM selection in an online setting, where the learner interacts with users through multi-step query refinement and must choose LLMs sequentially without access to offline datasets or model internals. A key challenge arises from unstructured context evolution: the prompt dynamically changes in response to previous model outputs via a black-box process, which cannot be simulated, modeled, or learned. To address this, we propose the first contextual bandit framework for sequential LLM selection under unstructured prompt dynamics. We formalize a notion of myopic regret and develop a LinUCB-based algorithm that provably achieves sublinear regret without relying on future context prediction. We further introduce budget-aware and positionally-aware (favoring early-stage satisfaction) extensions to accommodate variable query costs and user preferences for early high-quality responses. Our algorithms are theoretically grounded and require no offline fine-tuning or dataset-specific training. Experiments on diverse benchmarks demonstrate that our methods outperform existing LLM routing strategies in both accuracy and cost-efficiency, validating the power of contextual bandits for real-time, adaptive LLM selection.
Code is available at \url{https://github.com/EntroShape/Online_LLM_Selection}
\end{abstract}
\section{Introduction}\label{sec:intro}

Large Language Models (LLMs) have demonstrated remarkable capabilities across a wide range of tasks, from open-ended dialogue to complex reasoning and mathematical problem solving \cite{brown2020language, bubeck2023sparks, dai2025multi}. However, these capabilities come with significant trade-offs: more powerful models tend to be slower and more expensive, while faster, cheaper models often lack robustness or reasoning ability \cite{yuan2025selfrewardinglanguagemodels, anil2023palm}. In practice, no single model is universally optimal across all insputs or user preferences \cite{chen2025optimizing}.

This growing heterogeneity has motivated the deployment of \emph{multi-LLM systems}, where multiple LLMs with different strengths (e.g., accuracy, latency, cost) are available, and a routing or selection mechanism determines which one to use for a given user query. Such multi-model setups are increasingly adopted in both academic research and industrial applications, where balancing performance and efficiency is critical \cite{nguyen2024metallm, agrawal2023openai}. A lightweight model like Mistral may suffice for factual lookups or shallow reasoning, while more complex tasks might require high-end models like GPT-4 \cite{openai2023gpt4}. Selecting the right LLM at the right time can yield substantial improvements in cost-effectiveness and user satisfaction \cite{chen2025optimizing, dai2024cost}.

Existing approaches to multi-LLM routing  \cite{nguyen2024metallm, dai2024cost,aggarwal2024automix, wang2025mixllm} often rely on static policies, supervised learning on offline datasets, or single-step decision-making. However, real-world user interactions are often \emph{sequential and adaptive}. Users refine their queries based on prior answers, leading to multi-step interactions where the system can revise its LLM choices dynamically. For example, a user might ask a factual question, follow up with a clarification request, and then pose a deeper reasoning challenge. Each step introduces new contextual information—typically derived from the model’s earlier responses—which should influence subsequent model selection.

To support this type of dynamic, multi-step interaction, we propose a new framework for \emph{online multi-LLM selection under unstructured context evolution}. In our setting, the system interacts with a user over a sequence of steps: at each step, it receives a prompt context—potentially incorporating previous model responses—and must select an LLM to generate the next reply. Crucially, the prompt evolves over time according to a black-box transformation that may include concatenation, rewriting, user edits, or proprietary formatting. Because this evolution process is unknown, non-differentiable, and potentially stochastic, conventional planning-based approaches such as reinforcement learning (RL) are ill-suited. In effect, modeling how a prompt evolves would itself require reasoning at the level of an LLM, rendering the transition dynamics unlearnable in practice.

To address these challenges—multi-LLM decision-making, the need for online adaptation, and unstructured context evolution—we adopt a contextual bandit framework. Rather than modeling full interaction trajectories, we decompose the process into a sequence of \emph{myopic decisions}, where the system selects the best LLM for the current context without attempting to forecast how the prompt will evolve. This formulation is both feasible and effective: it allows us to sidestep modeling the unknown transition dynamics while still enabling continual improvement from user feedback. We formalize this learning objective using \emph{myopic regret}, which quantifies the loss relative to the best possible LLM choice at each step.
Building on this formulation, we develop a LinUCB-based algorithm that maintains confidence-aware estimates of each model's performance across varying prompt contexts. To handle practical deployment constraints, we further extend the algorithm to account for stochastic per-query costs and budget constraints, and introduce a heuristic that prioritizes high-quality responses early in the interaction to reflect user positional bias. Together, our framework offers a principled and scalable online solution for adaptive, cost-aware multi-LLM selection.

\textbf{Our Contributions.}
We make the following key contributions:
\begin{itemize}
    \item We introduce the first contextual bandit framework for multi-step LLM selection under \emph{unstructured context evolution}—a realistic and underexplored setting that captures the adaptive nature of interactive user–LLM interactions.
    \item We propose a Greedy LinUCB algorithm for per-step LLM selection and establish a sublinear regret bound with respect to \emph{myopic regret}. Our results show that efficient online learning is achievable even without modeling how prompts evolve over time.
    \item We extend our method to handle \emph{stochastic cost constraints} and \emph{position-sensitive utility}, designing budget-aware and position-aware variants that improve decision quality in cost-limited settings and prioritize early user satisfaction in multi-turn interactions.
    \item We conduct extensive experiments across diverse benchmarks, demonstrating that our algorithms consistently outperform existing routing methods in both accuracy and cost-efficiency. The results further confirm the value of context evolution and the effectiveness of lightweight online adaptation.
\end{itemize}

Overall, our work provides a principled and lightweight foundation for orchestrating multiple LLMs in open-ended, multi-step environments. By using contextual bandits under unstructured prompt evolution, we demonstrate that practical and scalable multi-LLM systems can be built without relying on offline training or complex environment simulation.

\textbf{Related Work.}

Several recent works~\cite{nguyen2024metallm,wang2025mixllm,xia2024llm} have studied LLM selection in multi-model deployments to improve cost-efficiency and performance. MetaLLM \cite{nguyen2024metallm} applies LinUCB to choose an LLM for each query based on learned reward signals derived from offline preference data. MixLLM \cite{wang2025mixllm} formulates LLM routing as a linear contextual bandit problem, optimizing a weighted sum of estimated accuracy, cost, and latency. While both methods leverage contextual bandits \cite{lattimore2020bandit,dai2024conversational}, they operate in a single-step setting and do not account for prompt evolution or multi-turn query refinement.
Other approaches rely on static or offline policies. AutoMix \cite{aggarwal2024automix} introduces a self-verification mechanism where a lightweight model decides whether to escalate a query to a stronger LLM, but assumes access to a reliable verification oracle and relies on a fixed cascade. Similarly, OpenAI’s Function Calling Selector \cite{agrawal2023openai} trains a static classifier to select between GPT variants based on supervised fine-tuning, without adapting to online feedback or context changes.
Additional recent efforts include C2MAB-V \cite{dai2024cost}, which adopts a combinatorial bandit approach for selecting LLM subsets under complex reward formulations, and LLMSelector \cite{chen2025optimizing}, which optimizes expert-model placement across distributed systems. The Multi-Armed Router \cite{zhang2024how} proposes an exploration–exploitation strategy for LLM routing but focuses on single-turn inference and assumes fixed prompt representations.
A growing body of work explores multi-turn LLM orchestration. ReAct \cite{yao2022react} and Tree of Thoughts \cite{yao2023tree} propose agent-style reasoning pipelines where LLMs operate over multiple steps using hand-crafted heuristics. These approaches demonstrate promising performance but rely on predefined planning rules and lack learnable routing policies. PromptAgent \cite{wang2023promptagent} automates prompt optimization using Monte Carlo Tree Search, but requires task-specific reward functions and simulated feedback, limiting its generality.

In contrast to all of the above, our work tackles the problem of online multi-step LLM selection under unstructured and unlearnable context evolution. We propose a contextual bandit formulation that is lightweight, feedback-driven, and equipped with provable regret guarantees—bridging the gap between practical black-box orchestration and theoretically grounded online learning.

\section{Motivation}\label{sec:motivation}

\begin{wrapfigure}{R}{7cm}
\centering
\centerline{\includegraphics[width=1\linewidth]{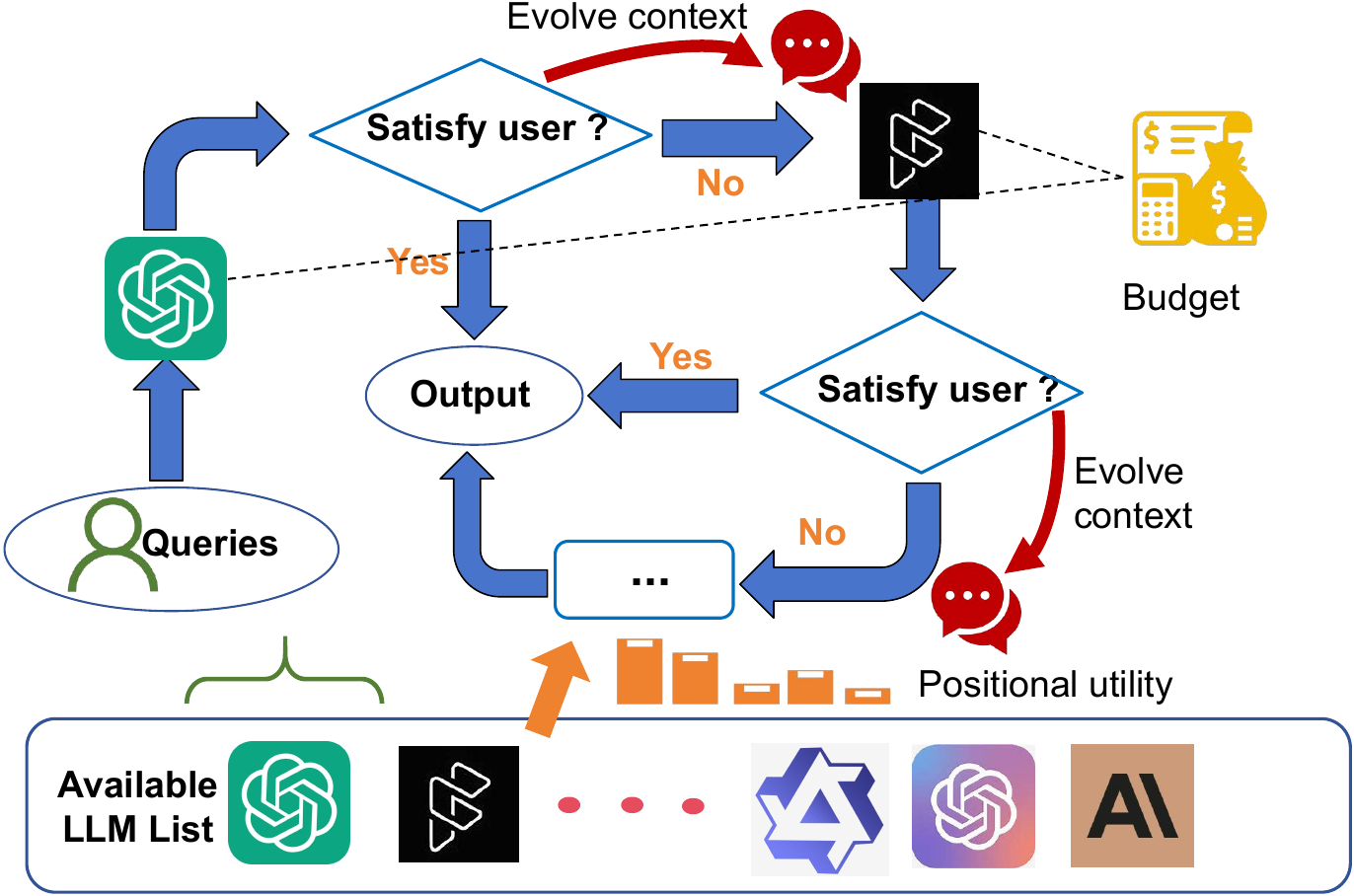}}
\caption{Online multi-LLM selection with contextual bandits under unstructured context evolution.}
\label{fig:overview}
\end{wrapfigure}

Deploying a single LLM to handle diverse user queries often results in unnecessary trade-offs between cost, latency, and performance. In this paper, we advocate for a sequential, multi-model approach to LLM orchestration—where a system adaptively selects among multiple LLMs over multiple steps in response to user feedback. In this section, we outline three motivating observations: the promise of multi-LLM selection, the importance and challenge of evolving context, and the suitability of contextual bandits.

\textbf{\emph{Why multi-LLM selection matters.}}
Modern LLMs exhibit diverse behaviors: they vary in factual accuracy, latency, and cost. As demonstrated in our experiments (Section ~\ref{sec:exp} ), no single model consistently dominates across all tasks. For instance, lightweight models like Mistral-Small-3.1 respond quickly and perform well on simple queries, while larger models like GPT-4.1 are better suited for complex reasoning. This motivates the use of \emph{multi-LLM selection}—dynamically choosing which LLM to use for each query—to achieve better performance-cost trade-offs.

This strategy is increasingly adopted in production systems that serve diverse user queries. Recent works and industry practices have explored LLM routing and orchestration to reduce latency and API costs while maintaining output quality (e.g., \cite{nguyen2024metallm, agrawal2023openai, aggarwal2024automix}). In such systems, multiple LLMs are deployed as a shared model pool, and a routing policy determines which one to invoke. Using a high-cost model like GPT-4.1 indiscriminately is often unnecessary and inefficient—especially for trivial or repetitive queries. A well-designed selector can route easy queries to cheaper models while reserving more capable models for harder cases, yielding significant cost savings without sacrificing quality.

\textbf{\emph{Why context matters and why it is unstructured.}}
A key design choice in our setting is to let each LLM selection depend not only on the original user query, but also on the responses of previously selected models. This results in an \emph{evolving prompt context}, where the input to each LLM is dynamically constructed based on prior outputs. Through empirical observations (see Section ~\ref{sec:exp}), we find that incorporating this context is crucial—later LLMs often perform better when they can ``see'' earlier attempts, as they may correct previous errors or build upon partial answers. This motivates our formulation of sequential multi-LLM selection with contextual adaptation.

However, modeling how the prompt evolves across steps is inherently difficult. The new context may be formed through a combination of LLM responses, prompt concatenation, user edits, or system-specific transformations. We formalize this process as a context evolution function $g$, which maps the current context and LLM output to the next-step prompt. Importantly, $g$ is a \emph{black-box function}—often non-differentiable, stochastic, or proprietary—and cannot be explicitly modeled or learned. In practice, predicting the next prompt would require simulating how an LLM response changes the prompt structure, which itself may require reasoning as complex as what the LLM performs. 
This complexity makes multi-step planning intractable and violates the assumptions of RL, which depends on having a learnable or simulatable transition model. By contrast, our framework avoids this issue by adopting a myopic view: it selects the best LLM for the \emph{current context}, without trying to predict future prompt evolution. This enables robust online decision-making even under unstructured and unpredictable context dynamics.

\textbf{\emph{Why contextual bandits are the right tool.}}
Given the limitations of RL in this setting, we turn to contextual bandits as a lightweight yet principled alternative. Rather than modeling full trajectories, a contextual bandit approach selects the best LLM for each step based solely on the current prompt context. This \emph{myopic decision-making} is both feasible and robust: it avoids the need to simulate or model $g$, while still adapting to evolving feedback over time.
As illustrated in Figure~\ref{fig:overview}, our framework supports multi-step interaction with users: at each step, the system observes a new prompt context, selects an LLM, receives feedback, and optionally evolves the context for the next step. We formalize the learning objective via \emph{myopic regret}, which compares the agent’s chosen LLM to the best possible LLM for the current prompt context. This yields a theoretically tractable and practically meaningful measure of performance, even when full trajectory optimization is infeasible.

Contextual bandits are particularly attractive in this setting because they are lightweight, data-efficient, and robust to dynamic environments \cite{dai2024online,abbasi2011improved,zuo2022hierarchical}. Unlike RL, they require no offline training or simulator access, and they adapt easily to new user populations, model APIs, and cost structures. This makes them well-suited for real-world multi-LLM systems, where queries are diverse, budgets vary, and models evolve rapidly. By combining theoretical rigor with practical feasibility, contextual bandits provide a compelling foundation for adaptive LLM routing in open-ended, real-time environments.

\section{Problem Formulation}\label{sec:model}
We consider an online sequential decision-making problem for adaptive multi-LLM selection, motivated by real-time user interactions with LLMs. Let $[K] = \{1, \dots, K\}$ denote the set of available LLMs. Each LLM $k \in [K]$ is associated with an unknown feature vector $\theta_k^* \in \mathbb{R}^d$, capturing its alignment with user preferences across different prompt contexts.
A learning agent interacts with users over $T$ rounds. In each round $t \in [T]$, the agent receives a user query $Q_t$ and engages in a multi-step adaptive interaction with the user, lasting at most $H$ steps. Let $x_{t,1} \in \mathbb{R}^d$ denote the initial context vector derived from $Q_t$ (e.g., an embedding or prompt representation).
At each step $h \in [H]$, the agent observes context $x_{t,h}$ and selects an LLM $a_{t,h} \in [K]$ to generate a response. Let $R_{t,h}$ denote the output of LLM $a_{t,h}$ when invoked on context $x_{t,h}$. 
The user provides binary feedback $r_{t,h} \in \{0,1\}$, indicating satisfaction with the result. For analytical simplicity, we model the feedback as a linear function of the context and the selected LLM’s parameters corrupted by noise:
\[
r_{t,h} = \langle x_{t,h}, \theta_{a_{t,h}}^* \rangle + \epsilon_{t,h},
\]
where $\epsilon_{t,h}$ is a zero-mean sub-Gaussian noise term. This surrogate model enables the application of linear contextual bandit algorithms despite the discrete nature of the actual user feedback.
If $r_{t,h} = 1$, the round ends. Otherwise, the agent proceeds to the next step, where the context is updated by incorporating the previous LLM response:
\begin{equation*}
    x_{t,h+1} = g(x_{t,h}, a_{t,h}, R_{t,h}, r_{t,h}),
\end{equation*}
where $g$ is a black-box function representing how the context evolves in response to the LLM's output (e.g., through concatenation, rewriting, or augmentation).
Importantly, as discussed in \Cref{sec:motivation}, $g$ is assumed to be unstructured and unlearnable: it may be stochastic, nonstationary, or governed by proprietary model behavior, and cannot be explicitly modeled or simulated. 
This interaction continues for up to $H$ steps or until the user is satisfied. The goal is to design an online decision policy that maximizes the expected cumulative user satisfaction over $T$ rounds.

\textbf{Myopic regret.}
Due to the unstructured and unobservable nature of the context evolution function $g$, multi-step planning or reinforcement learning methods that rely on modeling or simulating future states are inapplicable. Instead, we adopt a myopic regret minimization framework, where the agent selects each LLM based solely on the current context, without modeling future transitions.
At step $h$ of round $t$, define the instantaneous regret as:
\begin{equation*}
    \text{Reg}_{t,h} = \max_{k \in [K]} \langle x_{t,h}, \theta_k^* \rangle - \langle x_{t,h}, \theta_{a_{t,h}}^* \rangle,
\end{equation*}
which quantifies the expected shortfall in user satisfaction due to a suboptimal LLM choice in the given context.
The total regret over $T$ rounds is:
\begin{equation*}
\mathcal{R}(T) = \sum_{t=1}^T \sum_{h=1}^{H_t} \text{Reg}_{t,h},
\end{equation*}
where $H_t \le H$ denotes the number of steps taken in round $t$.

\emph{Remark.}
While it may seem natural to compare against the best full sequence of LLMs per query (i.e., sequence-wise regret), such analysis requires modeling or learning the transition function $g$, which is infeasible in practice. Moreover, the space of possible LLM trajectories is combinatorially large and highly nonstationary. Thus, sequence-wise regret is neither practically meaningful nor theoretically tractable in our setting. 
By contrast, myopic regret minimization ensures robust performance at each decision point, without relying on assumptions about future context evolution, making it more suitable for real-time multi-LLM decision-making under unstructured prompt dynamics

\section{Greedy LinUCB Algorithm}\label{sec:alg-greedy}

In this section, we propose a LinUCB-based greedy algorithm for sequential LLM selection. As discussed in \Cref{sec:model}, since future contexts evolve via an unknown and unstructured function $g$, multi-step planning is intractable. We therefore adopt a myopic strategy that focuses on optimizing per-step feedback using contextual bandits. Specifically, we adapt the LinUCB algorithm to this setting by maintaining a separate linear model for each LLM and selecting the arm with the highest upper confidence bound (UCB) at each step.

\begin{algorithm}[t]
\caption{Greedy LinUCB for Multi-LLM Selection}
\label{alg:greedy-linucb}
\begin{algorithmic}[1]
\STATE \textbf{Input:} Regularization parameter $\lambda > 0$, confidence parameter $\alpha > 0$
\STATE Initialize $A_k \leftarrow \lambda I_d$, $b_k \leftarrow \mathbf{0} \in \mathbb{R}^d$ for all $k \in [K]$
\FOR{round $t = 1, 2, \cdots$}
    \STATE Receive initial query $Q_t$ and build context $x_{t,1} \in \mathbb{R}^d$
    \FOR{step $h = 1$ to $H$}
        \FOR{each $k \in [K]$}
            \STATE $\hat{\theta}_k \leftarrow A_k^{-1} b_k$; \quad $\text{UCB}_k \leftarrow \langle x_{t,h}, \hat{\theta}_k \rangle + \alpha \cdot \sqrt{x_{t,h}^\top A_k^{-1} x_{t,h}}$
        \ENDFOR
        \STATE Query LLM $a_{t,h} \leftarrow \arg\max_k \text{UCB}_k$ with $(Q_t, R_{t,1}, \cdots, R_{t,h-1})$
        \STATE Receive output $R_{t,h}$ and binary feedback $r_{t,h} \in \{0,1\}$
        \STATE Update: $A_{a_{t,h}} \leftarrow A_{a_{t,h}} + x_{t,h} x_{t,h}^\top; \quad
        b_{a_{t,h}} \leftarrow b_{a_{t,h}} + r_{t,h} x_{t,h}$
        \IF{$r_{t,h} = 1$}
            \STATE \textbf{break}
        \ELSE
            \STATE $x_{t,h+1} \leftarrow g(x_{t,h}, a_{t,h}, R_{t,h}, r_{t,h})$
        \ENDIF
    \ENDFOR
\ENDFOR
\end{algorithmic}
\end{algorithm}

As described in \Cref{alg:greedy-linucb}, the greedy LinUCB algorithm maintains a regularized least-squares model for each LLM $k \in [K]$, represented by a matrix $A_k$ and a response vector $b_k$. At each decision step $(t,h)$, the algorithm computes the ridge regression estimate $\hat{\theta}_k = A_k^{-1} b_k$ and evaluates the LinUCB index for each LLM (line 7). 
This index combines the predicted reward $\hat{\theta}k^\top x{t,h}$ with an exploration bonus proportional to the estimate's uncertainty, promoting the selection of under-explored LLMs. The LLM with the highest LinUCB score is chosen for querying. This approach is crucial, as selecting solely based on the highest predicted reward (pure exploitation) risks suboptimal decisions by neglecting parameter uncertainty.
After receiving the model response $R_{t,h}$ and observing the binary user feedback $r_{t,h} \in \{0,1\}$, the algorithm updates the corresponding model parameters $A_{a_{t,h}}$ and $b_{a_{t,h}}$. If the user is satisfied (i.e., $r_{t,h} = 1$), the round terminates. Otherwise, the context evolves to $x_{t,h+1} = g(x_{t,h}, a_{t,h}, R_{t,h}, r_{t,h})$, and the algorithm proceeds to the next step. This greedy procedure is repeated at each decision point without modeling the future evolution of context.

Under the following assumptions, we provide a high-probability upper bound on the cumulative myopic regret of Algorithm~\ref{alg:greedy-linucb}.

\begin{assumption}[Bounded parameters]
For all $k \in [K]$, $\|\theta_k^*\|_2 \le S$.
\end{assumption}

\begin{assumption}[Bounded context norm]
For all $x_{t,h}$, it holds that $\|x_{t,h}\|_2 \le L$.
\end{assumption}

\begin{assumption}[Linear feedback]
The expected feedback satisfies $\mathbb{E}[r_{t,h}] = \langle x_{t,h}, \theta_{a_{t,h}}^* \rangle$.
\end{assumption}

\begin{assumption}[Sub-Gaussian noise]
The noise $\epsilon_{t,h}$ is conditionally zero-mean and sub-Gaussian.
\end{assumption}

\begin{theorem}
\label{thm:greedy-linucb}
Under Assumptions 1–4, with probability at least $1 - \delta$, the cumulative myopic regret of the Greedy LinUCB algorithm satisfies:
\[
\mathcal{R}(T) = O\left( \sqrt{K d T H} \cdot \left( S L + \sqrt{\lambda} S \right) \cdot \log\left( \frac{K T L^2}{\lambda \delta} \right) \right).
\]
\end{theorem}

The regret bound in \Cref{thm:greedy-linucb} matches the structure of standard LinUCB bounds in contextual bandits, up to an additional factor of $\sqrt{H}$ reflecting the multi-step interaction in each round. The regret scales as $\widetilde{O}(S L \sqrt{K d T H})$, where the $\widetilde{O}$ notation hides logarithmic terms. Compared to classical contextual bandits \cite{lattimore2020bandit,abbasi2011improved}, the challenge in our setting stems from the evolving and unmodeled context dynamics. Nevertheless, by minimizing per-step regret and avoiding assumptions on the transition function $g$, the greedy LinUCB algorithm remains provably efficient and robust in this unstructured sequential environment.

\section{Budget-Aware Algorithms under Stochastic Costs}

In real-time LLM selection settings, different models incur different costs due to latency, compute, or token-based pricing. Moreover, these costs can vary stochastically with user queries. Motivated by this, we extend our algorithm to handle per-query \emph{random costs}, where each user specifies a budget and the learner must select LLMs adaptively within that budget while maximizing user satisfaction.

\paragraph{Problem setup.} In each round $t \in [T]$, the user provides a total budget $B_t > 0$. At each step $h \in [H]$, when the learner selects arm $a_{t,h}$, it observes a reward $r_{t,h} \in \{0,1\}$ and incurs a cost $c_{t,h,a_{t,h}} \in [0, C_{\max}]$ drawn i.i.d. from a fixed but unknown distribution with mean $\mu_{a_{t,h}}$. The learner observes the cost only after querying the LLM. The cumulative cost must satisfy $
\sum_{h=1}^{H_t} c_{t,h,a_{t,h}} \le B_t$, where $H_t$ is the number of steps taken in round $t$. The goal is to maximize cumulative user satisfaction under this per-round budget constraint.

\subsection{Budget-Aware Greedy LinUCB Algorithm.} 

At each step, the algorithm computes:
\begin{itemize}
    \item \textbf{Reward estimate:} $\text{UCB}_k(x_{t,h}) = \langle x_{t,h}, \hat{\theta}_k \rangle + \alpha_k$, where $\hat{\theta}_k$ is the LinUCB estimator and $\alpha_k$ is the confidence width.
    \item \textbf{Cost estimate:} $\hat{c}_k$ is the empirical mean of observed costs for arm $k$, with confidence interval width $\beta_k = \sqrt{\log(2TK/\delta) / (2N_k)}$, where $N_k$ is the number of times arm $k$ has been selected.
    \item \textbf{Score:} The algorithm computes
    \[
    \text{Score}_k(x_{t,h}) = \frac{\text{UCB}_k(x_{t,h})}{\max\{\hat{c}_k - \beta_k, \varepsilon\}},
    \]
   where $\varepsilon$ is a small positive constant to avoid a denominator of 0, and selects the arm with the highest score whose upper-bound cost estimate $\hat{c}_k + \beta_k$ fits within the remaining budget.
    
\end{itemize}

This two-level confidence mechanism ensures both optimism in reward and conservatism in cost, balancing exploration and budget feasibility. We now provide a regret bound comparing against a per-step myopic oracle that knows true reward and cost expectations.

\paragraph{Myopic regret definition.} In the presence of budget constraints and stochastic costs, it is infeasible to compare against a full-horizon optimal policy or to use standard cumulative reward as the regret benchmark. Instead, we adopt a \emph{per-step myopic regret} formulation. At each step $(t,h)$, define the best feasible arm (oracle) as:
\[
k^*_{t,h} = \arg\max_{k \in [K],\ \mu_k \le b_{t,h}} \frac{\langle x_{t,h}, \theta_k^* \rangle}{\mu_k},
\]
where $b_{t,h}$ is the remaining budget before step $h$ in round $t$. This oracle greedily selects the arm with the highest expected reward per unit cost that fits within the remaining budget. Let the algorithm’s selected arm be $a_{t,h}$. Then the per-step regret is:
\[
\text{Reg}_{t,h} = \langle x_{t,h}, \theta_{k^*_{t,h}}^* \rangle - \langle x_{t,h}, \theta_{a_{t,h}}^* \rangle,
\]
and the total regret is:
\[
\mathcal{R}_{\text{budget}}(T) = \sum_{t=1}^T \sum_{h=1}^{H_t} \text{Reg}_{t,h}.
\]

\begin{assumption}[Stochastic Cost]
For each arm $k \in [K]$, the cost $c_{t,h,k}$ is i.i.d. with unknown mean $\mu_k \in (0, C_{\max}]$ and sub-Gaussian noise.
\end{assumption}

\begin{theorem}
\label{thm:stochastic-cost-regret}
Under Assumptions 1–5 and with probability at least $1 - \delta$, the cumulative myopic regret of budget-aware greedy LinUCB  algorithm satisfies
\[
\mathcal{R}_{\text{budget}}(T) = \widetilde{O}\left( S L \cdot \sqrt{K d T H} + \sum_{k=1}^K \frac{C_{\max}}{\mu_k^2} \cdot \sqrt{T \log\left(\frac{TK}{\delta}\right)} \right).
\]
\end{theorem}

This result extends classical LinUCB analysis to the setting with budget constraints and stochastic, unknown costs. The reward regret scales with $\sqrt{K d T H}$, while the additional cost regret grows with $\sqrt{T}$ and is amplified by small mean costs $\mu_k$. Despite this challenge, the algorithm remains budget-feasible with high probability and achieves sublinear cumulative regret.

\subsection{Positionally-Aware Knapsack Heuristic}

The greedy budget-aware algorithm introduced earlier selects LLMs sequentially by maximizing a cost-normalized reward score. While effective in balancing reward and cost, this approach may overly prioritize cheap LLMs and delay the use of more capable ones until later in the interaction. In many real-world scenarios, however, users exhibit positional bias: they prefer to receive high-quality responses early and may disengage if early attempts fail. This motivates the need to consider \emph{positional utility}, where earlier successful responses are more valuable in practice than later ones.

\begin{algorithm}[t]
\caption{Positionally-Aware Knapsack Heuristic}
\label{alg:knapsack-heuristic}
\begin{algorithmic}[1]
\REQUIRE Budget $B_t$, UCB scores $\widehat{r}_k$, cost estimates $\hat{c}_k$ for all $k \in [K]$
\STATE Initialize remaining budget $b \leftarrow B_t$ and candidate list $\mathcal{C} \leftarrow \emptyset$
\WHILE{$b > 0$}
    \STATE Solve a 0-1 knapsack problem on arms $[K] \setminus \mathcal{C}$ with budget $b$ and objective $(\widehat{r}_k, \hat{c}_k)$
    \STATE Let $k^\text{next} \leftarrow$ arm with highest $\widehat{r}_k$ among selected items in knapsack solution
    \IF{$k^\text{next}$ does not exist or $\hat{c}_{k^\text{next}} > b$}
        \STATE \textbf{break}
    \ENDIF
    \STATE Add $k^\text{next}$ to candidate list $\mathcal{C}$
    \STATE $b \leftarrow b - \hat{c}_{k^\text{next}}$
\ENDWHILE
\STATE \textbf{return} candidate list $\mathcal{C}$
\end{algorithmic}
\end{algorithm}

To better balance early accuracy and overall cost, we propose a positionally-aware knapsack heuristic in \Cref{alg:knapsack-heuristic}. The key idea is to incorporate cost estimates and UCB reward scores into a dynamic selection procedure that prioritizes early deployment of high-confidence LLMs while respecting the per-query budget. At each step, the algorithm solves a 0-1 knapsack problem over the remaining LLM candidates, using UCB scores as reward estimates and empirical costs as weights. From the knapsack solution, it selects the arm with the highest UCB score, adds it to the candidate list, and updates the remaining budget. This process is repeated until no feasible arms remain or the budget is exhausted. By anchoring each decision within a budget-aware optimization, the method ensures that strong LLMs are selected as early as possible without exceeding cost constraints.

\emph{Remark.} This knapsack-based strategy explicitly targets positional utility by favoring the early use of high-performing models, rather than deferring them due to their cost. Unlike purely greedy cost-normalized approaches, which may over-prioritize cheap but weak models, the knapsack heuristic balances exploration and resource allocation more effectively. While the method is heuristic and lacks a formal regret bound, it aligns closely with practical deployment goals in interactive systems. As demonstrated in our experiments in , the algorithm delivers higher effective utility—especially when stronger models are costly and positional preferences matter.
\section{Experiments}\label{sec:exp}
\noindent\textbf{Baselines.} We compare our algorithms against online LLM routing methods.
For online, contextual bandit approaches, MetaLLM \cite{nguyen2024metallm} employs LinUCB to select an LLM by learning a policy that predicts a reward (balancing accuracy and cost) based on query embeddings.
MixLLM \cite{wang2025mixllm}, another online bandit method, routes queries by optimizing a weighted sum of predicted response quality, cost, and an explicit latency penalty. Reported optimal trade-off parameter $\lambda = 1.4$ is used for MixLLM. Majority Voting~\cite{li2024more} is also considered as a baseline due to its effectiveness and simplicity.

\noindent\textbf{Candidate LLMs.} The LLM pool, used across all baseline comparisons, included several state-of-the-art models (i) \textit{Deepseek-v3}, a state-of-the-art open-source MoE model from China with 671 billion total parameters; (ii) \textit{llama-4-maverick}, a latest open-source Mixture-of-Experts (MoE) model with 400 billion total parameters and 128 experts;  (iii) \textit{phi-4}, a 14-billion-parameter model from Microsoft, designed for tasks including complex reasoning and math problem-solving; (iv) \textit{gemini-2.0-flash}, a multimodal model from Google DeepMind, suitable for rapid responses. (v) \textit{Mistral-small-3.1-24b-instruct}, an efficient instruction-following model from Mistral AI; (vi) \textit{GPT-4.1-Nano}, a lightweight variant of GPT-4.1, optimized for fast and cost-effective responses.

\noindent\textbf{Datasets.} Experiments were conducted on various datasets covering diverse tasks and domains. (i) \textit{MMLU-Pro}~\cite{wang2024mmlu}: A comprehensive multiple-choice question (MCQ) benchmark covering topics across Science, Technology, Engineering, and Mathematics (STEM) disciplines. (ii) \textit{Math-500}~\cite{lightman2023lets}: A question-answering (QA) benchmark featuring mathematical problems across seven distinct undergraduate-level subjects. (iii) \textit{GPQA}~\cite{rein2024gpqa}: A dataset consisting of challenging graduate-level multiple-choice questions (MCQs) in physics, biology, and chemistry. For our experiments, we converted the MCQs into open-ended questions to further increase the difficulty. (iv) \textit{AIME}: A QA benchmark consisting of problems from the American Invitational Mathematics Examination, testing advanced mathematical reasoning. Our dataset comprised of the data from AIME\footnote{\url{https://huggingface.co/datasets/MathArena/aime_2025}} 2024 and 2025\footnote{\url{https://huggingface.co/datasets/Maxwell-Jia/AIME_2024}}.

\noindent\textbf{Evaluation Protocol.} For \textit{MMLU-Pro}, accuracy was used as the evaluation metric. For the remaining datasets, responses were evaluated using \textit{Deepseek-R1}\cite{liu2024deepseek} as an LLM grader that provided binary feedback (``true'' or ``false'') based on whether the generated answer matched the gold answer. These feedback signals were then used to compute accuracy scores. Each dataset was randomly split into an offline training set (20\%) for initialization and an online testing set (80\%) for evaluation.

\noindent\textbf{Experiment Specification.} For our proposed algorithms, we set a maximum cascade depth of $H=4$ LLMs per query. Query embeddings were generated using the sentence~\cite{bge_embedding}. Query embeddings for all evaluated algorithms are in 384-dimensional embeddings. The exploration parameter ($\alpha$) was set to 0.675 and the regularization parameter ($\lambda$) to 0.45. The given budget for each query for both budget-aware algorithms is randomly +/- 5\% to the average cost per query of greedy LinUCB. No any cost constarint was implemented For Greedy LinUCB algorithm.

\subsection{Experimental Results}

\subsubsection{Main Results}

\begin{table}[H]
\centering
\small
\caption{Accuracy (\%) on the different task. The best is in \textbf{bold} and the worst is \underline{underlined}. Random routing is not considered for best/worst.}
\label{tbl:main_final} 
\begin{tabular}{clcccc|c}
\toprule
&& MMLU-Pro & AIME & GPQA & Math500 & Avg \\
\midrule
\multirow{6}{*}{\rotatebox{90}{\textit{\small Candidate LLMs}}}
& Mistral-Small-3.1 & 48.80 & \underline{1.67} & \underline{22.22} & \underline{57.60} & \underline{32.57} \\
& Phi-4 & 51.50 & 8.33 & 29.80 & 67.20 & 39.21 \\
& Llama-4-Maverick & 41.77 & \textbf{20.00} & \textbf{39.90} & 85.40 & 46.77 \\
& Gemini-2.0-Flash & \textbf{62.10} & \textbf{20.00} & 35.30 & \textbf{86.00} & \textbf{50.85} \\
& GPT-4.1-Nano & \underline{41.33} & 6.67 & 29.80 & 71.60 & 37.35 \\
& DeepSeek-V3-Chat & 58.80 & 3.33 & 31.31 & 70.40 & 40.96 \\
\midrule
& Voting & 63.30 & 67.39 & 47.48 & 50.85 & 57.26 \\ 
\midrule
\multirow{6}{*}{\rotatebox{90}{\textit{\small Routing}}}
& MixLLM & 76.32 & 41.67 & 53.79 & 85.20 & 64.25 \\ 
& MetaLLM & \underline{72.11} & \underline{20.83} & 52.40 & \underline{83.80} & \underline{57.29} \\ 
& Random & 49.75 & 13.64 & 27.85 & 67.00 & 39.56 \\
&  Greedy LinUCB & \textbf{86.25} & 54.16 & \underline{51.28} & \textbf{96.25} & 71.9 \\
&  Budget Aware LinUCB & 77.29 & 56.25 & 59.12 & 84.00 & 69.17 \\ 
&  Positionally-Aware Knapsack & 82.20 & \textbf{62.50} & \textbf{66.67} & 88 & \textbf{74.84} \\ 
\bottomrule
\end{tabular}
\end{table}

\begin{table}[H]
\centering
\small
\caption{Cost (US Dollars). The best (lowest) is in \textbf{bold} and the worst (highest) is \underline{underlined}. Random routing is not considered for best/worst.}
\label{tbl:cost}
\begin{tabular}{clrrrr|c}
    \toprule
    & & MMLU-Pro & AIME & GPQA & Math500 & Avg \\
    \midrule
    \multirow{6}{*}{\rotatebox{90}{\textit{\small Candidate LLMs}}}
    & Mistral-Small-3.1 & \textbf{2.00E-05} & 3.72E-03 & \underline{1.08E-02} & 5.44E-05 & \underline{3.65E-03} \\
    & Phi-4 & \textbf{2.00E-05} & 3.82E-03 & \textbf{5.05E-05} & \textbf{4.83E-05} & 9.85E-04 \\
    & Llama-4-Maverick & 8.30E-05 & \textbf{1.41E-04} & 1.34E-04 & 1.02E-04 & \textbf{1.15E-04} \\
    & Gemini-2.0-Flash & 2.80E-05 & 3.01E-04 & 1.06E-04 & \underline{2.07E-04} & 1.61E-04 \\
    & GPT-4.1-Nano & 2.70E-05 & \underline{1.19E-02} & 1.20E-04 & 1.31E-04 & 3.04E-03 \\
    & DeepSeek-V3-Chat & \underline{1.16E-04} & 2.37E-04 & 1.85E-04 & 1.62E-04 & 1.75E-04 \\
    \midrule
    & Voting & {3.38E-04} & {2.31E-02} & {1.31E-02} & {8.10E-04} & {9.34E-03} \\
    \midrule
    \multirow{6}{*}{\rotatebox{90}{\textit{\small Routing}}}
    & MixLLLM & \underline{1.27E-04} & \underline{3.61E-03} & \underline{9.40E-03} & \underline{4.77E-04} & \underline{3.40E-03} \\
    & MetaLM & 6.73E-05 & \textbf{4.26E-04} & 5.24E-04 & 3.28E-04 & 3.36E-04 \\
    & Random & 4.73E-05 & 1.56E-04 & 9.81E-05 & 1.17E-04 & 1.05E-04 \\
    & Greedy LinUCB & 9.28E-05 & 2.14E-03 & 9.80E-04 & 4.31E-04 & 9.11E-04 \\
    & Budget-Aware LinUCB & \textbf{4.89E-05} & 7.00E-04 & \textbf{3.40E-04} & 9.72E-05 & \textbf{2.97E-04} \\
    & Positionally-Aware Knapsack Heuristic & 5.12E-05 & 7.52E-04 & 5.10E-03 & \textbf{9.57E-05} & 1.50E-03 \\
    \bottomrule
\end{tabular}
\end{table}

The experimental results demonstrate the effectiveness of the proposed online multi-LLM selection algorithms.
As seen from (\Cref{tbl:main_final}), the Positionally-Aware Knapsack heuristic achieved the highest average accuracy of 74.84\%. This represents a 16.5\% improvement over the best-performing baseline router, MixLLM (64.25\%), and is also notably higher than our other proposed methods, Greedy LinUCB (71.99\%) and Budget-Aware LinUCB (69.17\%). In terms of cost (\Cref{tbl:cost}), this heuristic (1.50E-03 USD average cost) is approximately 2.27 times more economical than MixLLM (3.32E-03 USD average cost) and substantially cheaper than the Voting \cite{li2024more} baseline (9.34E-03 USD average cost).

The Budget-Aware LinUCB algorithm stands out for its cost-efficiency, with an average cost of 2.97E-04 USD (\cref{tbl:cost}). This is comparable to the cost of the MetaLLM baseline (3.36E-04 USD average cost) and approaches the cost of the most economical individual LLM, Llama-4-Maverick (1.15E-04 USD average cost). However, Budget-Aware LinUCB delivers a significantly higher average accuracy of 69.17\% (\cref{tbl:main_final}). This is 20.7\% greater than MetaLLM (57.71\% average accuracy) and 39.6\% higher than the most accurate individual LLM, Gemini-2.0-Flash (50.85\% average accuracy), while being over 3.07 times more economical than our Greedy LinUCB (8.29E-04 USD average cost) for a similar accuracy. Even compared to the Random routing baseline, which has a low average cost (1.05E-04 USD), Budget-Aware LinUCB achieves 79.4\% higher average accuracy (69.17\% vs. 39.56\%) for a moderately higher, yet still very low, operational cost.

The restricted average accuracy of baseline methods, MixLLM (64.25\%) and MetaLLM (57.71\%), might be attributed to their formulation of optimizing a weighted sum of estimated accuracy, cost, and latency (MixLLM) or learned reward signals (MetaLLM) for each query independently. Both approaches operate within a single-step decision-making paradigm; their routing logic is applied only once per user query based on the initial context. This design inherently lacks the capability to adapt to or leverage the context evolution that occurs in multi-turn user interactions, where subsequent prompts are dynamically modified based on prior LLM outputs. Their myopic optimization, focused solely on the current step's predicted outcome, limits their effectiveness in optimizing performance over an entire multi-step conversation sequence, which is a core characteristic of real-world user-LLM interactions and our experimental setup.

\subsubsection{Effectiveness of Positional Awareness}

\begin{table}[htbp] 
\centering
\caption{Accuracy (\%) and Average Steps by Position for Different Configurations.}
\label{tbl:positionaccuracy_reordered} 
\resizebox{\columnwidth}{!}{
\begin{NiceTabular}{clrrrr|r} 
    \toprule
    Config. & Position & MMLU-Pro & AIME & GPQA & Math500 & Avg \\
    \midrule
    \Block{6-1}{Greedy LinUCB} 
    & Total Accuracy  & 86.25 & 54.16 & 51.28 & 96.25 & 71.99 \\
    & Avg. Step & 1.870 & 3.375 & 3.308 & 1.31 & 2.52  \\
    \cline{2-7} 
    & 1st      & 58.58 & 18.75 & 28.21 & 86.00 & 47.89 \\
    & 2nd      & 15.04 & 20.83 & 15.38 & 6.25 & 15.38 \\
    & 3rd      & 7.12  & 10.42 & 2.56  & 3.00  & 6.03  \\
    & 4th      & 5.05  & 4.17  & 5.13  & 1.00  & 4.09  \\
    
    \midrule
    \Block{6-1}{Budget-Aware Greedy LinUCB} 
    & Overall  & 77.29 & 56.25 & 59.12 & 84 & 69.17 \\
    & Avg Step & 2.120 & 2.042 & 2.897 & 1.08 & 2.04  \\
    \cline{2-7} 
    & 1st      & 48.21 & 47.92 & 32.70 & 80.75 & 52.39 \\
    & 2nd      & 14.62 & 4.17  & 11.95 & 3.00  & 8.435  \\
    & 3rd      & 9.12  & 2.08  & 8.18  & 0.25  & 4.90  \\
    & 4th      & 5.33  & 2.08  & 6.29  & 0  & 3.49  \\
    \midrule
    \Block{6-1}{Positionally-Aware Knapsack heuristic} 
    & Overall  & 82.20 & 62.50 & 66.67 & 88.00 & 74.84 \\
    & Avg Step & 1.996 & 1.542 & 1.780 & 1.0075 & 1.58  \\
    \cline{2-7} 
    & 1st      & 78.11 & 60.42 & 57.86 & 88.00 & 71.10 \\
    & 2nd      & 4.05  & 2.08  & 7.55  & 0.00  & 3.42  \\
    & 3rd      & 0.04  & 0.00  & 1.26  & 0.00  & 0.33  \\
    & 4th      & 0.00  & 0.00  & 0.00  & 0.00  & 0.00  \\
    \bottomrule
\end{NiceTabular}} 
\end{table}

The Positionally-Aware Knapsack heuristic is specifically designed to prioritize early user satisfaction, reflecting a positional discount where earlier high-quality responses are more valuable. \cref{tbl:positionaccuracy_reordered} illustrates this design's success. The Knapsack heuristic achieves an average accuracy of 71.10\% in the very first step. This remarkable initial performance accounts for approximately 95.0\%  of its total average accuracy of 74.84\%. This indicates that for the vast majority of queries, the Knapsack heuristic identifies a satisfactory LLM on its first attempt. In contrast, the other proposed methods, while effective, distribute their accuracy accumulation over more steps. The Greedy LinUCB method, for instance, achieves 47.89\% accuracy in the first step, which is only about 71.24\%  of its 71.99\% overall average accuracy. Similarly, the Budget-Aware Greedy LinUCB obtains 52.39\% accuracy in its initial attempt, constituting roughly 75.7\% of its total average accuracy of 69.17\%. Moreover, the Knapsack heuristic satisfies user queries in an average of just 1.58 steps, significantly fewer than the 2.52 steps for Greedy LinUCB and 2.04 steps for Budget-Aware Greedy LinUCB. These results underscore the effectiveness of solving a knapsack problem at each step to select high-confidence LLMs early while respecting budget constraints. This approach aligns well with practical needs in multi-turn interactions, where users often value prompt and accurate responses, thereby enhancing the overall user experience.

\subsubsection{Effectiveness of Budget Awareness}

\begin{figure}[H]
    \centering
    
    \begin{subfigure}[b]{0.48\textwidth} 
        \centering
        \includegraphics[width=\linewidth]{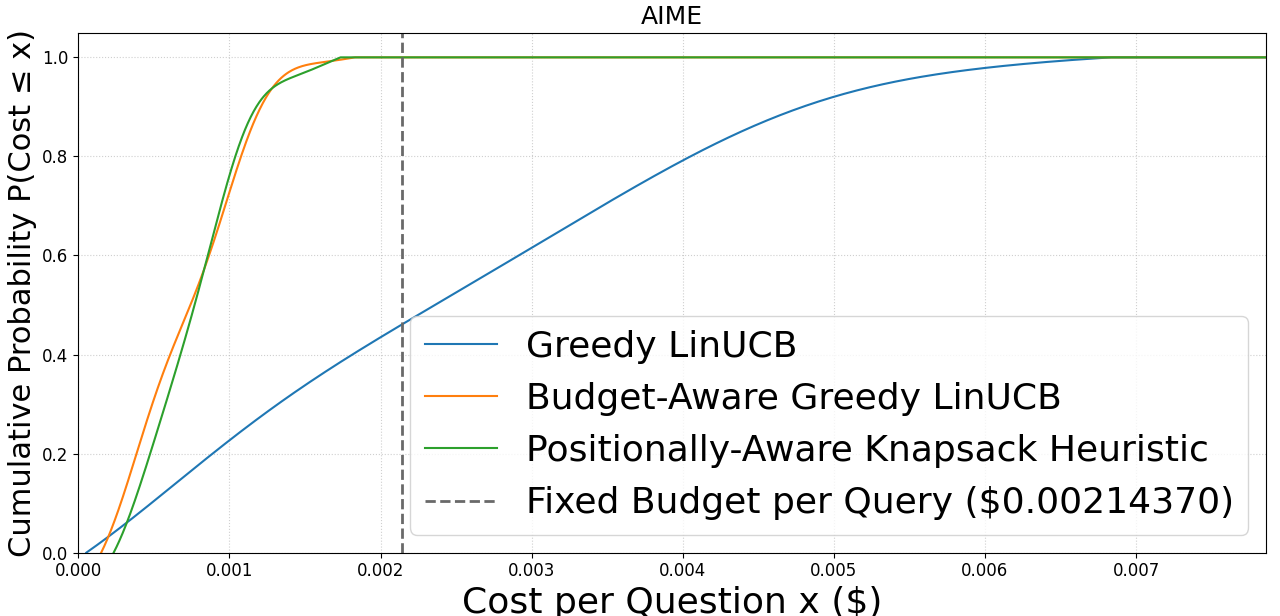} 
        \caption{AIME}
        \label{fig:graph1} 
    \end{subfigure}
    \hfill
    \begin{subfigure}[b]{0.48\textwidth}
        \centering
        \includegraphics[width=\linewidth]{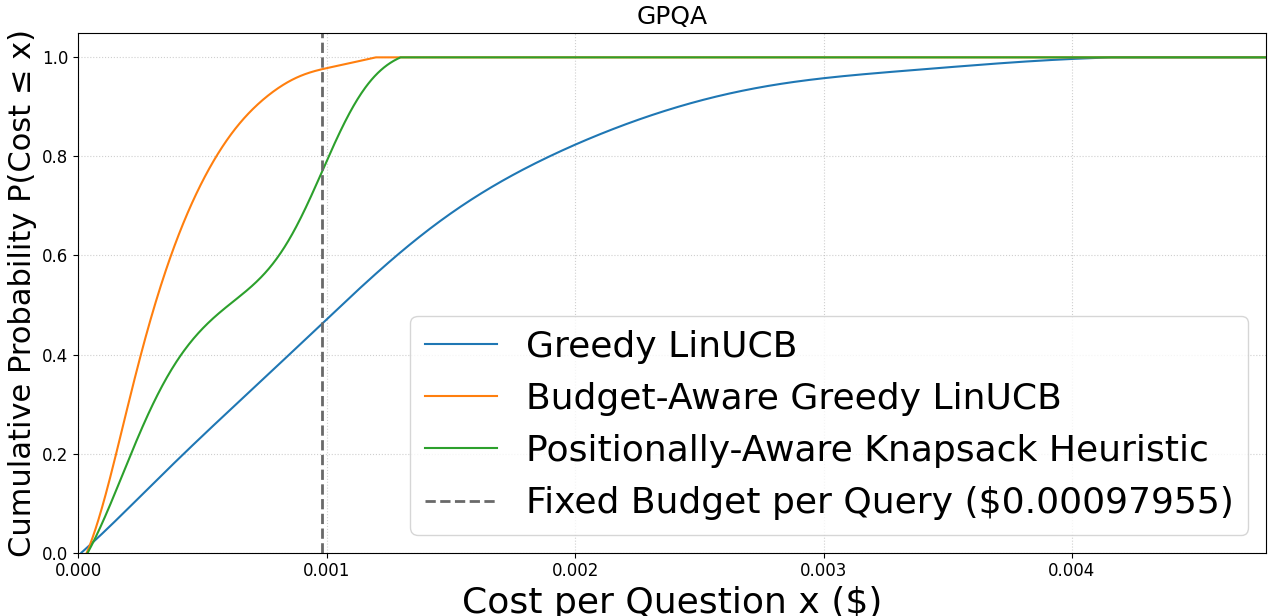} 
        \caption{GPQA}
        \label{fig:graph2} 
    \end{subfigure}

    \vspace{0.5cm}
    
    \begin{subfigure}[b]{0.48\textwidth}
        \centering
        \includegraphics[width=\linewidth]{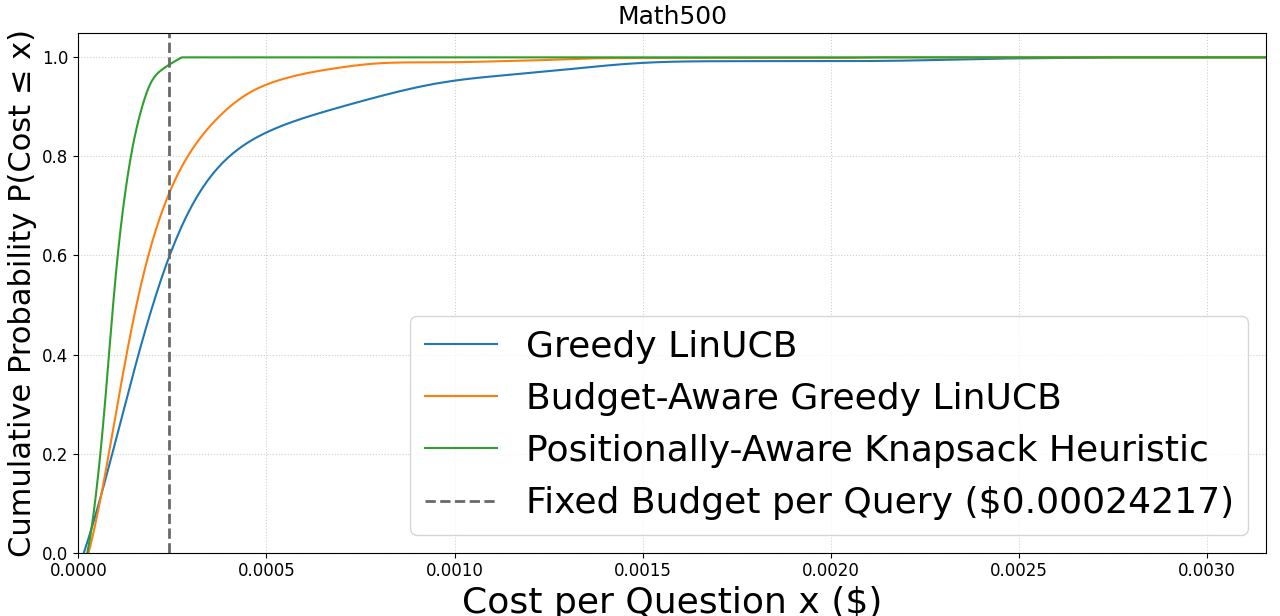}
        \caption{Math500}
        \label{fig:graph3} 
    \end{subfigure}
    \hfill 
    \begin{subfigure}[b]{0.48\textwidth}
        \centering
        \includegraphics[width=\linewidth]{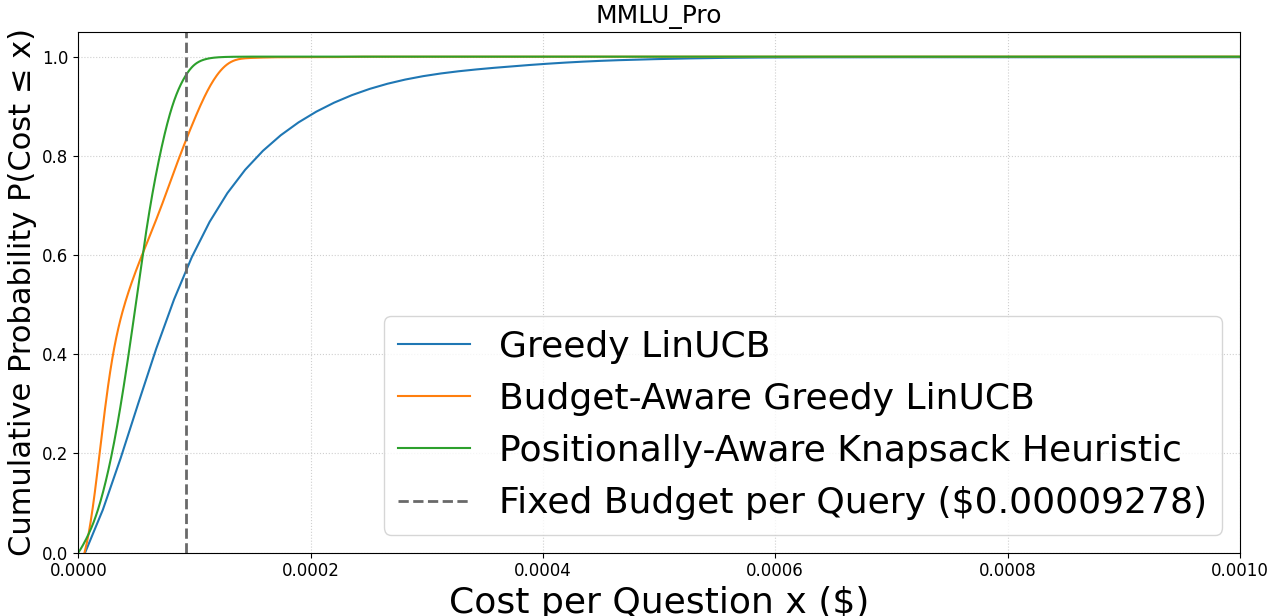}
        \caption{MMLU\_Pro}
        \label{fig:graph4} 
    \end{subfigure}

    \caption{CDF of per-query cost.}
    \label{fig:all_graphs} 
\end{figure}
\vspace{0.2cm}

The practical deployment of multi-LLM systems necessitates robust cost control, a challenge directly addressed by our budget-aware algorithms. Figure~\ref{fig:all_graphs} presents Cumulative Distribution Function (CDF) plots of per-query costs, where curves terminating at or before the designated fixed budget line (dashed vertical line) signify effective budget adherence. Across all datasets, the Budget-Aware Greedy LinUCB algorithm consistently demonstrates superior cost management; its CDF curve typically saturates near or before the budget threshold, indicating that the vast majority of queries are processed within the allocated financial limits. This performance contrasts sharply with the standard Greedy LinUCB whose cost distribution frequently extends well beyond the budget line, reflecting its lack of explicit cost constraints and resulting in higher average operational expenses as noted in \Cref{tbl:cost}. The Positionally-Aware Knapsack heuristic, while primarily optimizing for early user satisfaction, also exhibits commendable budget discipline due to its cost-aware knapsack formulation, generally keeping query costs within bounds. These distributions affirm that incorporating explicit budget-aware mechanisms, as in Budget-aware LinUCB, is highly effective and crucial for enabling cost-efficient LLM selection, ensuring that the system can reliably operate within predefined financial constraints without unduly sacrificing performance.

\subsubsection{Sensitivity Analysis on Budget Constraint}

\begin{figure}
\centering
\centerline{\includegraphics[width=1\linewidth]{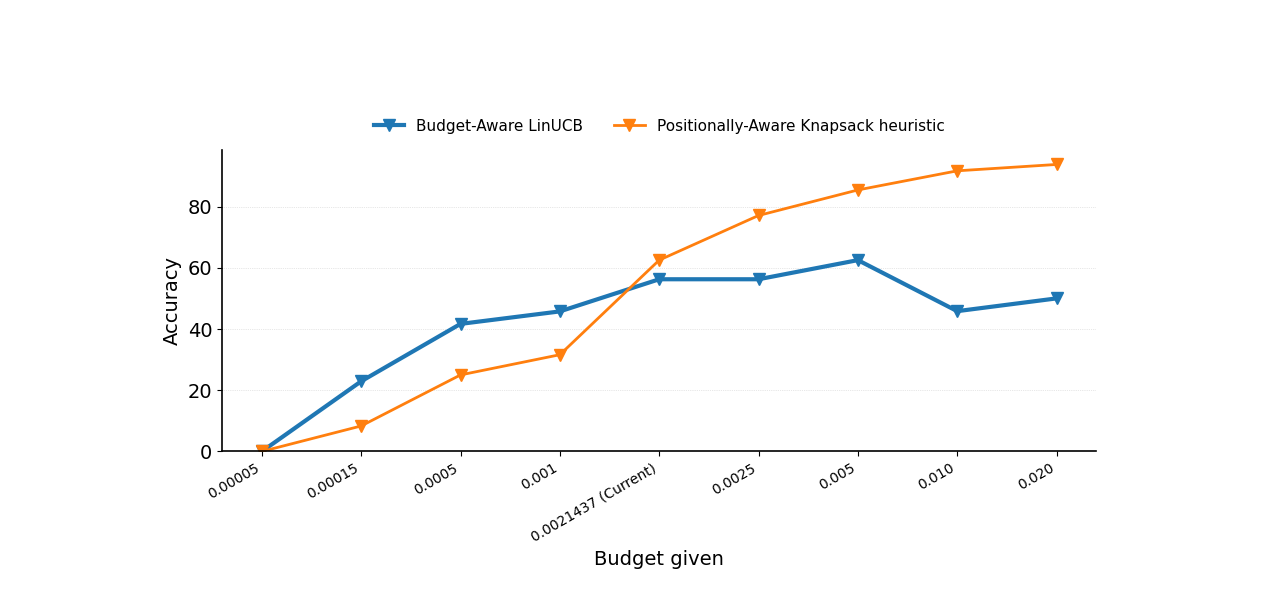}}
\caption{Sensitivity Analysis on Budget Constraint on AIME Dataset.}
\end{figure}

To assess the impact of varying budget constraints, we evaluated the performance of Budget-Aware LinUCB and the Positionally-Aware Knapsack heuristic across a range of fixed budgets on the AIME Dataset. As can be seen in Figure~\ref{fig:all_graphs} , at extremely low budgets (e.g., 0.00005), both algorithms achieve zero performance, as the budget is likely insufficient to select any LLM. As the budget increases, the performance of both algorithms generally improves. Budget-Aware LinUCB initially outperforms the Positionally-Aware Knapsack heuristic at tighter budget constraints (e.g., from 0.00015 to 0.001). However, as the budget increases further, the Positionally-Aware Knapsack heuristic demonstrates superior performance, surpassing Budget-Aware LinUCB around the ``Current'' operational budget (0.0021437) and achieving significantly higher performance at larger budgets (0.0035 to 0.02). Notably, the Knapsack heuristic's performance consistently scales with increasing budget, reaching its peak at the highest tested budget (0.02). In contrast, Budget-Aware LinUCB's performance peaks around a budget of 0.005 and subsequently declines at higher budget allocations. This suggests that while Budget-Aware LinUCB is effective under strict budget limitations, the Positionally-Aware Knapsack heuristic offers greater performance scalability and robustness as budget availability increases, better leveraging larger budgets for higher overall utility.

\section{Concluding Remarks}
This work presents a contextual bandit framework for online multi-LLM selection under unstructured prompt evolution, offering a lightweight and theoretically grounded solution. While our algorithms demonstrate strong performance across diverse benchmarks, several limitations remain. First, we adopt a myopic decision strategy and do not model long-term interaction effects, which may limit performance in tasks requiring deep multi-step reasoning. Second, the framework assumes access to reliable user feedback, which may be noisy or delayed in practice. Third, we treat the LLM pool as fixed, whereas practical deployments may involve dynamic model availability or changing costs. Future work includes extending the framework to handle delayed and noisy feedback, incorporating non-myopic planning under partially learnable prompt dynamics, and optimizing LLM scheduling in dynamically evolving model pools.
\newpage
\bibliographystyle{unsrt}
\bibliography{reference}
\newpage
\appendix
\section*{Appendix}
\section{Proofs}

\subsection{Proof of Theorem~\ref{thm:greedy-linucb}}
\begin{proof}
We analyze the cumulative myopic regret of the Greedy LinUCB algorithm under a linear reward model: for each arm $k \in [K]$ and context $x_{t,h}$, the expected reward is $\mathbb{E}[r_{t,h}] = \langle x_{t,h}, \theta_k^* \rangle$.

Let $A_k = \lambda I + \sum x x^\top$ and $b_k = \sum r x$ be the design matrix and response vector constructed from past observations for arm $k$, and define the ridge regression estimate $\hat{\theta}_k = A_k^{-1} b_k$.

By Theorem 2 of \cite{abbasi2011improved}, under Assumptions 1–4, with probability at least $1 - \delta$, for all arms $k$ and all contexts $x_{t,h}$:
\[
|\langle \hat{\theta}_k - \theta_k^*, x_{t,h} \rangle| \le \alpha \cdot \|x_{t,h}\|_{A_k^{-1}},
\]
where
\[
\alpha = O\left( (S L + \sqrt{\lambda} S) \cdot \sqrt{\log\left( \frac{K T L^2}{\lambda \delta} \right)} \right).
\]

Let $k^*_{t,h} = \arg\max_k \langle x_{t,h}, \theta_k^* \rangle$ be the optimal arm at step $(t,h)$, and let $a_{t,h}$ be the arm selected by Greedy LinUCB. The instantaneous regret is:
\[
\text{Reg}_{t,h} = \langle x_{t,h}, \theta_{k^*_{t,h}}^* \rangle - \langle x_{t,h}, \theta_{a_{t,h}}^* \rangle.
\]
Using the definition of the UCB index and the confidence bounds:
\[
\text{Reg}_{t,h} \le 2 \alpha \cdot \|x_{t,h}\|_{A_{a_{t,h}}^{-1}}.
\]

Summing over all $T$ rounds and up to $H$ steps per round:
\[
\mathcal{R}(T) = \sum_{t=1}^T \sum_{h=1}^{H_t} \text{Reg}_{t,h}
\le 2 \alpha \sum_{t=1}^T \sum_{h=1}^{H_t} \|x_{t,h}\|_{A_{a_{t,h}}^{-1}}.
\]

Using the standard cumulative uncertainty bound (Lemma 11 of \cite{abbasi2011improved}):
\[
\sum_{t,h} \|x_{t,h}\|_{A_{a_{t,h}}^{-1}}^2 \le K d \log\left(1 + \frac{T H L^2}{\lambda d}\right),
\]
and applying Cauchy–Schwarz:
\[
\sum_{t,h} \|x_{t,h}\|_{A_{a_{t,h}}^{-1}} \le \sqrt{T H K d \cdot \log\left(1 + \frac{T H L^2}{\lambda d}\right)}.
\]

Substituting into the regret bound:
\[
\mathcal{R}(T) = O\left( \alpha \cdot \sqrt{K d T H} \cdot \sqrt{\log\left( \frac{T H L^2}{\lambda d} \right)} \right).
\]

Combining logarithmic terms and substituting the expression for $\alpha$, we conclude:
\[
\mathcal{R}(T) = O\left( \sqrt{K d T H} \cdot \left( S L + \sqrt{\lambda} S \right) \cdot \log\left( \frac{K T L^2}{\lambda \delta} \right) \right). \qedhere
\]
\end{proof}

\subsection{Proof of Theorem~\ref{thm:stochastic-cost-regret}}
\begin{proof}
We decompose the total regret into two parts: (i) regret from suboptimal reward estimation, assuming true costs are known, and (ii) additional regret due to cost estimation errors.

\textbf{Total Regret Decomposition.}
At each step $(t,h)$, let the myopic oracle select
\[
k^*_{t,h} = \arg\max_{k \in [K],\ \mu_k \le b_{t,h}} \frac{\langle x_{t,h}, \theta_k^* \rangle}{\mu_k},
\]
and let $a_{t,h}$ be the algorithm’s selected arm. Then the per-step regret is
\[
\text{Reg}_{t,h} = \frac{\langle x_{t,h}, \theta_{k^*_{t,h}}^* \rangle}{\mu_{k^*_{t,h}}} \cdot \mu_{k^*_{t,h}} - \frac{\langle x_{t,h}, \theta_{a_{t,h}}^* \rangle}{\mu_{a_{t,h}}} \cdot \mu_{a_{t,h}}.
\]

This can be decomposed as:
\[
\text{Reg}_{t,h} = \left( \frac{\langle x_{t,h}, \theta_{k^*_{t,h}}^* \rangle}{\mu_{k^*_{t,h}}} - \frac{\langle x_{t,h}, \theta_{a_{t,h}}^* \rangle}{\mu_{a_{t,h}}} \right) \cdot \mu_{k^*_{t,h}} + \langle x_{t,h}, \theta_{a_{t,h}}^* \rangle \cdot \left( \frac{\mu_{k^*_{t,h}} - \mu_{a_{t,h}}}{\mu_{a_{t,h}}} \right).
\]

For analysis, we upper bound this regret by two terms:
\[
\text{Reg}_{t,h} \le \underbrace{\left| \frac{\langle x_{t,h}, \theta_{k^*_{t,h}}^* \rangle}{\mu_{k^*_{t,h}}} - \frac{\langle x_{t,h}, \theta_{a_{t,h}}^* \rangle}{\mu_{a_{t,h}}} \right|}_{\text{(A): reward estimation error under true costs}} \cdot \mu_{k^*_{t,h}} + \underbrace{\left| \frac{\mu_{k^*_{t,h}} - \mu_{a_{t,h}}}{\mu_{a_{t,h}}} \right|}_{\text{(B): cost estimation error}}.
\]

\textbf{(A) Reward regret.}
If the cost estimates $\mu_k$ were known, the algorithm reduces to a cost-weighted Greedy LinUCB variant. As in the standard LinUCB analysis, the cumulative regret from reward estimation over $T$ rounds and $H$ steps is:
\[
\mathcal{R}_{\text{reward}}(T) = \widetilde{O}\left( S L \cdot \sqrt{K d T H} \right).
\]

\textbf{(B) Cost regret.}
The algorithm uses an empirical estimate $\hat{c}_k$ and confidence width $\beta_k$ to compute the denominator in the score:
\[
\text{Score}_k(x_{t,h}) = \frac{\text{UCB}_k(x_{t,h})}{\hat{c}_k - \beta_k}.
\]
With high probability, Hoeffding’s inequality ensures:
\[
|\hat{c}_k - \mu_k| \le \beta_k = \sqrt{\frac{\log(2TK/\delta)}{2N_k}}.
\]

This affects the ranking of arms. In the worst case, a suboptimal arm $k$ with a slightly underestimated cost may be chosen over the optimal one, even if its reward is lower. Over $T$ rounds, each such deviation contributes an additional regret. Assuming the maximum deviation in cost estimate affects selection for $O(\sqrt{T})$ rounds and the worst-case reward difference is $O(1)$, we obtain:
\[
\mathcal{R}_{\text{cost}}(T) = \widetilde{O}\left( \sum_{k=1}^K \frac{C_{\max}}{\mu_k^2} \cdot \sqrt{T \log\left( \frac{TK}{\delta} \right)} \right),
\]
where the factor $\frac{1}{\mu_k^2}$ arises from the sensitivity of the score to cost estimation errors near small $\mu_k$.

\textbf{Total regret.}
Adding the two components:
\[
\mathcal{R}_{\text{budget}}(T) = \mathcal{R}_{\text{reward}}(T) + \mathcal{R}_{\text{cost}}(T)
= \widetilde{O}\left( S L \cdot \sqrt{K d T H} + \sum_{k=1}^K \frac{C_{\max}}{\mu_k^2} \cdot \sqrt{T \log\left(\frac{TK}{\delta}\right)} \right). \qedhere
\]
\end{proof}

\section{Impact of Context}

Our framework posits that leveraging evolving context---the output from previous LLM interactions---can enhance the performance of subsequent LLM selections. To specifically investigate this, we conducted a controlled experiment on 60 questions from the AIME dataset, comparing two strategies: Strategy 1, using Gemini-Flash as a standalone solver, and Strategy 2, a two-step sequential approach where Mistral-Small first attempts the problem, and its output (regardless of correctness) is then provided as context to Gemini-Flash for a second attempt. One example for positive and negative context impact is provided in appendix. 

The results demonstrate a tangible benefit from context. Strategy 2 (Mistral $\rightarrow$ Gemini w/ Context) achieved a 25\% success rate, outperforming Strategy 1's (Gemini-Only) 20\% success rate. More tellingly, in 8 out of the 60 questions, Gemini-Only failed, Mistral's initial attempt also failed, but Gemini with context from Mistral's attempt subsequently succeeded. This directly indicates that the contextual information, even from an initially incorrect response, can provide valuable grounding, expose flawed reasoning paths to avoid, or offer a different starting point that enables the subsequent LLM to reach the correct solution. This aligns with the core motivation of our work: that adaptive, multi-step interactions where context evolves can lead to better outcomes than single-shot LLM applications. 

However, the influence of context is not universally positive. In 3 instances, the Gemini-Only strategy succeeded, while the sequential Mistral $\rightarrow$ Gemini approach failed. This suggests that poor or misleading initial context can sometimes derail a subsequent LLM that might have otherwise succeeded. Such cases underscore the complexity of context interaction and highlight a potential challenge in multi-LLM cascades. The example below demonstrates a scenario where the initial context might have had a detrimental effect.

Overall, while the sequential strategy incurred a higher average cost per question (\$0.000712 for Strategy 2 vs. \$0.000229 for Strategy 1) due to invoking two models in many cases, the net gain in accuracy (5\% percentage points, with 8 direct context-driven successes) supports the hypothesis that context evolution is a valuable mechanism. This targeted experiment highlights the potential of contextual information to improve problem-solving in multi-LLM systems, providing a foundational rationale for the adaptive, context-aware selection mechanisms developed in this paper. The instances of negative impact also motivate further research into robustly determining when and how context should be passed or filtered. Below we have provided two examples that show the impact of context. 

\begin{tcolorbox}[
  enhanced,                     
  colback=White,               
  colframe=DarkSlateGray,     
  title=Example : Context Helped,              
  fonttitle=\bfseries,          
  coltitle=White,              
  colbacktitle=DarkSlateGray,  
  attach boxed title to top left={ 
    yshift=-2mm, 
    xshift=2mm   
  },
  breakable                   
]

\textbf{Question:}  Six points $A, B, C, D, E$ and $F$ lie in a straight line in that order. Suppose that $G$ is a point not on the line and that $AC = 26$, $BD = 22$, $CE = 31$, $DF = 33$, $AF = 73$, $CG = 40$, and $DG = 30$. Find the area of $\triangle BGE$.
Ground Truth: 468

\textbf{Ground Truth: } 468

\vspace{\baselineskip}

\textbf{Strategy 1 (Gemini-Only):} Let $AB = a, BC = b, CD = c, DE = d, EF = e$.
$AC = a+b = 26, BD = b+c = 22, CE = c+d = 31, DF = d+e = 33, AF = a+b+c+d+e = 73$.
$AF = AC+CE+EF - CE + DF = 26+31+e = 73$, implying $a = 18$ and $e = 17$. $b = 8$, $c = 14$, $d = 17$. Therefore $AE = 51$.
$BE = BC + CE = 8 + 31 = 39$. Let G be $(0, y)$. Calculate B and E using distance formula.
Coordinates $B=(x_B, 0), E = (x_E, 0), G = (x_G, 0)$. Area of $\triangle BGE = \frac12|x_B-x_E| \cdot y$. Using law of cosines to coordinate points is too complex. Let the area of the triangle be denoted by $K$. No easy spatial representation.

The final answer is \boxed{195}

\textbf{Is Correct:} False

\textbf{Total Cost:} \$0.000155

Strategy 2 (Mistral -> Gemini w/ Context):

  Attempt 1 (Mistral):
    Model: mistralai/mistral-small-3.1-24b-instruct
    
    Parsed Answer: Given the points \(A, B, C, D, E, F\) on a line and \(G\) not on the line, we need to find the area of \(\triangle BGE\).     

1. Extend \(G\) to meet the line at point \(H\) such that \(GH \perp EF\).
2. Let \(EH = x\) and \(HF = y\).
3. According to Pythagoras' theorem, for \(\triangle GIH\): \(40^2 - (y+31)^2 = x^2\).
4. For \(\triangle GJH\): \(30^2 - (y+9)^2 = (x+31)^2\).
5. Solving these equations:
   \[
   1600 - (y+31)^2 = x^2
   \]
   \[
   900 - (y+9)^2 = (x+31)^2
   \]

6. Simplifying and solving these, coupled with equations found from \(73^2 - (x+y)^2 = (x+31)^2 + (y+9)^2\), leads to:
   \[
   [EH] = 16.0, [HF] = 20.5
   \]

7. Using the area formula \( \frac{1}{2} \times base \times height \):
   \[
   \text{Area of } \triangle BGE = \frac{1}{2} \times 44.5 \times 31 = 695.75
   \]

The final calculated area is \(\boxed{695.75}\).

\textbf{Is Correct:} False

  Attempt 2 (Gemini w/ Context):
    Model: google/gemini-2.0-flash-001
    
    Parsed Answer: Let $a, b, c, d, e, f$ be the positions of points $A, B, C, D, E, F$ on the line, respectively. We are given:
$c - a = 26$, $d - b = 22$, $e - c = 31$, $f - d = 33$, $f - a = 73$.
From $f - a = 73$, $f = a + 73$.
Also, $c = a + 26$, $d = b + 22$, $e = c + 31 = a + 26 + 31 = a + 57$, $f = d + 33 = b + 22 + 33 = b + 55$.
So $a + 73 = b + 55 \implies b = a + 18$. Then $d = a + 18 + 22 = a + 40$.
We have $DG = 30$ and $CG = 40$. Let $G = (x, y)$.
$(x - (a+26))^2 + y^2 = 40^2 = 1600$
$(x - (a+40))^2 + y^2 = 30^2 = 900$
Subtracting, $(x - a - 26)^2 - (x - a - 40)^2 = 700$.
$((x - a) - 26)^2 - ((x - a) - 40)^2 = 700$
$(x - a)^2 - 52(x - a) + 26^2 - [(x - a)^2 - 80(x - a) + 40^2] = 700$
$28(x - a) + 676 - 1600 = 700$
$28(x - a) = 700 + 924 = 1624$
$x - a = \frac{1624}{28} = 58$.
Thus $x = a + 58$.
Then $(a + 58 - a - 26)^2 + y^2 = 32^2 + y^2 = 1600$, so $y^2 = 1600 - 1024 = 576$, and $y = 24$.
Therefore, $B = a + 18$, $E = a + 57$, $G = (a + 58, 24)$.
Area of $\triangle BGE = \frac{1}{2} |(a + 18)(0 - 24) + (a + 57)(24 - 0) + (a + 58)(0 - 0)| = \frac{1}{2} |(18 - 57)(-24)| = \frac{1}{2} |(-39)(-24)| = \frac{1}{2} (39)(24) = 39 \cdot 12 = 468$.

The final answer is \boxed{468}

\textbf{Is Correct:} True
  
\textbf{Total Cost:} \$0.000504

\end{tcolorbox}

\begin{tcolorbox}[
  enhanced,                     
  colback=White,               
  colframe=DarkSlateGray,     
  title=Example : Context Hurted,              
  fonttitle=\bfseries,          
  coltitle=White,              
  colbacktitle=DarkSlateGray,  
  attach boxed title to top left={ 
    yshift=-2mm, 
    xshift=2mm   
  },
  breakable                   
]

\textbf{Question (AIME) :} Question: Let $O(0,0), A(\tfrac{1}{2}, 0),$ and $B(0, \tfrac{\sqrt{3}}{2})$ be points in the coordinate plane. Let $\mathcal{F}$ be the family of segments $\overline{PQ}$ of unit length lying in the first quadrant with $P$ on the $x$-axis and $Q$ on the $y$-axis. There is a unique point $C$ on $\overline{AB}$, distinct from $A$ and $B$, that does not belong to any segment from $\mathcal{F}$ other than $\overline{AB}$. Then $OC^2 = \tfrac{p}{q}$, where $p$ and $q$ are relatively prime positive integers. Find $p + q$.
\vspace{\baselineskip} 
\textbf{Ground Truth:} 23

\vspace{\baselineskip} 
\textbf{Strategy 1 (Gemini-Only):}Parsed Answer: Let $P(x,0)$ and $Q(0,y)$. Since $PQ=1$, $x^2+y^2=1$. The envelope of $PQ$ is $x^{2/3}+y^{2/3}=1$.
The coordinates of $C$ are where $x^{2/3}+y^{2/3}=1$ and $y=-\sqrt{3}x+\frac{\sqrt{3}}{2}$.
Solving, $C=(\frac{1}{8}, \frac{3\sqrt{3}}{8})$. Then $OC^2=\frac{1}{64}+\frac{27}{64}=\frac{28}{64}=\frac{7}{16}$. $p=7, q=16$. $p+q=23$.       

The final answer is \boxed{23}

  \textbf{Is Correct:} True
  
  \textbf{Total Cost for Attempt:} \$0.000114

\vspace{\baselineskip} 
\textbf{Strategy 2 (Mistral -> Gemini w/ Context):}

  Attempt 1 (Mistral):

    Model: mistralai/mistral-small-3.1-24b-instruct
    
    Parsed Answer: 1. **Define the points and the segment**:
   - Points are \( O(0,0) \), \( A\left(\frac{1}{2}, 0\right) \), and \( B\left(0, \frac{\sqrt{3}}{2}\right) \).
   - Segment \(\overline{PQ}\) has \( P \) on the \( x \)-axis and \( Q \) on the \( y \)-axis with unit length.

    2. **Parameterize point \( C \) on \( \overline{AB} \)**:
   - Let \( C \) be on \( \overline{AB} \) with coordinates \( C(x, y) \).
   - The line \( \overline{AB} \) can be parameterized as \( C(t) = (1-t)A + tB \) for \( 0 \leq t \leq 1 \).
   - Thus, \( C(t) = \left(\frac{1}{2}(1-t), \frac{\sqrt{3}}{2}t\right) \).

    3. **Find the condition for \( C \) not belonging to any segment in \( \mathcal{F} \)**:
   - For \( C \) to not belong to any segment in \( \mathcal{F} \), it must satisfy the condition where the distance from \( C \) to any point \( P \) on the \( x \)-axis and \( Q \) on the \( y \)-axis is greater than 1.
   - This implies \( \sqrt{x^2 + y^2} > 1 \).
   - Substituting \( x \) and \( y \) from the parameterization, we get:
     \[
     \sqrt{\left(\frac{1}{2}(1-t)\right)^2 + \left(\frac{\sqrt{3}}{2}t\right)^2} > 1
     \]
   - Simplifying inside the square root:
     \[
     \sqrt{\frac{1}{4}(1-t)^2 + \frac{3}{4}t^2} > 1
     \]
     \[
     \sqrt{\frac{1 - 2t + t^2 + 3t^2}{4}} > 1
     \]
     \[
     \sqrt{\frac{1 + 2t^2 - 2t}{4}} > 1
     \]
     \[
     \frac{\sqrt{1 + 2t^2 - 2t}}{2} > 1
     \]
     \[
     \sqrt{1 + 2t^2 - 2t} > 2
     \]
     \[
     1 + 2t^2 - 2t > 4
     \]
     \[
     2t^2 - 2t - 3 > 0
     \]
     \[
     t^2 - t - \frac{3}{2} > 0
     \]

4. **Solve the quadratic inequality**:
   - The roots of the quadratic equation \( t^2 - t - \frac{3}{2} = 0 \) are:
     \[
     t = \frac{1 \pm \sqrt{1 + 6}}{2} = \frac{1 \pm \sqrt{7}}{2}
     \]
   - The inequality \( t^2 - t - \frac{3}{2} > 0 \) holds for \( t < \frac{1 - \sqrt{7}}{2} \) or \( t > \frac{1 + \sqrt{7}}{2} \).

5. **Determine the unique \( t \) within the interval \( (0, 1) \)**:
   - Since \( 0 \leq t \leq 1 \), the valid \( t \) is \( t = \frac{1 + \sqrt{7}}{2} \).

6. **Calculate \( OC^2 \)**:
   - Substitute \( t = \frac{1 + \sqrt{7}}{2} \) back into the parameterization:
     \[
     C\left(\frac{1}{2}\left(1 - \frac{1 + \sqrt{7}}{2}\right), \frac{\sqrt{3}}{2}\left(\frac{1 + \sqrt{7}}{2}\right)\right)
     \]
     \[
     C\left(\frac{1}{2}\left(\frac{1 - 1 - \sqrt{7}}{2}\right), \frac{\sqrt{3}}{2}\left(\frac{1 + \sqrt{7}}{2}\right)\right)
     \]
     \[
     C\left(-\frac{\sqrt{7}}{4}, \frac{\sqrt{3}(1 + \sqrt{7})}{4}\right)
     \]
   - Calculate \( OC^2 \):
     \[
     OC^2 = \left(-\frac{\sqrt{7}}{4}\right)^2 + \left(\frac{\sqrt{3}(1 + \sqrt{7})}{4}\right)^2
     \]
     \[
     OC^2 = \frac{7}{16} + \frac{3(1 + 2\sqrt{7} + 7)}{16}
     \]
     \[
     OC^2 = \frac{7}{16} + \frac{3(8 + 2\sqrt{7})}{16}
     \]
     \[
     OC^2 = \frac{7}{16} + \frac{24 + 6\sqrt{7}}{16}
     \]
     \[
     OC^2 = \frac{31 + 6\sqrt{7}}{16}
     \]

7. **Simplify and find \( p \) and \( q \)**:
   - Since \( \sqrt{7} \) is irrational, we need to ensure \( p \) and \( q \) are relatively prime.
   - The expression simplifies to \( \frac{31 + 6\sqrt{7}}{16} \), but we need \( \frac{p}{q} \) to be rational.
   - Thus, we re-evaluate the steps and find the correct \( t \) that simplifies to a rational \( OC^2 \).

After re-evaluating, the correct \( t \) and simplifying, we find:

\[
OC^2 = \frac{19}{7}
\]

Thus, the final answer is:
\[
\boxed{26}
\]

    \textbf{Is Correct:} False
    
     \textbf{Total Cost for Attempt:} \$0.000155
    \vspace{\baselineskip} 
\textbf{Attempt 2 (Gemini w/ Context):
}   
 \textbf{Model: google/gemini-2.0-flash-001:} 
 
 Parsed Answer: Let $P=(x,0)$ and $Q=(0,y)$ with $x, y \ge 0$ and $PQ=1$. Then $x^2 + y^2 = 1$. The line PQ is given by $\frac{x'}{x} + \frac{y'}{y} = 1$. The equation of the line AB is $\frac{x'}{1/2} + \frac{y'}{\sqrt{3}/2} = 1$, or $2x' + \frac{2}{\sqrt{3}}y' = 1$.
Let $x=\cos\theta$ and $y=\sin\theta$ for $0\le \theta \le \pi/2$. Then the equation of the line PQ is $\frac{x'}{\cos\theta} + \frac{y'}{\sin\theta} = 1$, or $x' \sin\theta + y' \cos\theta = \sin\theta \cos\theta$.
The point $C(x',y')$ on AB is $2x' + \frac{2}{\sqrt{3}} y' = 1$.
We want to find a point $C$ on AB such that $x'\sin\theta + y' \cos\theta = \sin\theta \cos\theta$ has a unique solution for $\theta$.
Substituting $x' = \frac{1}{2} - \frac{1}{\sqrt{3}}y'$ gives
$(\frac{1}{2} - \frac{1}{\sqrt{3}}y')\sin\theta + y'\cos\theta = \sin\theta \cos\theta$
$\frac{1}{2}\sin\theta - \frac{1}{\sqrt{3}}y' \sin\theta + y'\cos\theta = \sin\theta \cos\theta$
$y'(\cos\theta - \frac{1}{\sqrt{3}} \sin\theta) = \sin\theta \cos\theta - \frac{1}{2} \sin\theta$
$y' = \frac{\sin\theta (\cos\theta - \frac{1}{2})}{\cos\theta - \frac{1}{\sqrt{3}} \sin\theta} = \frac{\sin\theta}{\cos\theta-\frac{1}{\sqrt{3}}\sin\theta} (\cos\theta - \frac{1}{2})$
We want $y'$ to have a unique solution. This arises when $\cos\theta - \frac{1}{\sqrt{3}}\sin\theta = 0$, i.e., $\tan\theta = \sqrt{3}$, so $\theta = \frac{\pi}{3}$.
In this case, $\cos\theta - \frac{1}{2} = 0$, so $\frac{1}{2} - \frac{1}{2} = 0$.
Consider $f(\theta) = \frac{\sin\theta}{\cos\theta - \frac{1}{\sqrt{3}}\sin\theta}$.
Let $\theta = \frac{\pi}{3}$ be the unique solution. Then $\cos\theta = \frac{1}{2}$ and $\sin\theta = \frac{\sqrt{3}}{2}$.
Then $x' = \frac{1}{2} - \frac{1}{\sqrt{3}}y'$.
So, $2x' + \frac{2}{\sqrt{3}}y' = 1$. The final answer is \boxed{7/16}

    \textbf{Is Correct:} False
    
    \textbf{Total Cost:}  \$0.000504

\end{tcolorbox}

\section{Prompt Template}

This section presents the prompts used to interact with the LLMs for both problem-solving and answer evaluation.

\begin{lstlisting}[
    basicstyle=\ttfamily\footnotesize,
    breaklines=true,
    caption={Base Prompt for Language Model Problem Solving.},
    label=lst:llm_base_prompt,
    frame=single, % Adds a frame around the listing
    numbers=left, % Adds line numbers
    numberstyle=\tiny\color{gray}, % Style for line numbers
    escapechar=` % Allows using ` to escape to LaTeX
]
Solve the following math problem: {question_text}

Please provide a Structured derivation of your answer within 50 words.
At the end, clearly state your final answer in LaTeX format, enclosed within `\textbackslash{}boxed`{}.
For example: 'The final answer is `\textbackslash{}boxed`{x=5}'.
\end{lstlisting}

\begin{lstlisting}[
    basicstyle=\ttfamily\footnotesize,
    breaklines=true,
    caption={Augmented Prompt for Language Model with Feedback from Prior Attempts.},
    label=lst:llm_augmented_prompt,
    frame=single,
    numbers=left,
    numberstyle=\tiny\color{gray},
    escapechar=`
]
Solve the following math problem: {question_text}

Please provide a Structured derivation of your answer within 50 words.
At the end, clearly state your final answer in LaTeX format, enclosed within `\textbackslash{}boxed`{}.
For example: 'The final answer is `\textbackslash{}boxed`{x=5}'.

Note: The following previous attempts were incorrect. Please provide a different solution or identify the error in previous reasoning if possible:
- Attempt 1 (by {failed_llm_id_1}) led to: {failed_answer_1}
- Attempt 2 (by {failed_llm_id_2}) led to: {failed_answer_2}
\end{lstlisting}

\begin{lstlisting}[
    basicstyle=\ttfamily\footnotesize,
    breaklines=true,
    caption={Prompt for the Grader Model to Evaluate LLM Answer Correctness.},
    label=lst:grader_prompt,
    frame=single,
    numbers=left,
    numberstyle=\tiny\color{gray},
    escapechar=`
]
Expression 1: {llm_answer_latex}
Expression 2: {ground_truth_latex}

Expression 2 is the answer and expression is attempt by student, look at their final answer only which might be boxed, does student get the final expected answer? Respond with only the word 'True' or 'False'.
\end{lstlisting}

 \end{document}